\documentclass[journal]{IEEEtran}
\usepackage{amsfonts,amssymb}
\usepackage{booktabs}

\usepackage{amsthm}
\usepackage{graphicx} 
\usepackage{amsmath}
\usepackage{amssymb}
\usepackage{subfigure}
\usepackage{epstopdf}
\usepackage{algorithm}
\usepackage{algpseudocode}
\usepackage{multirow}
\usepackage{array}

\usepackage{ragged2e}
\usepackage{hyperref}

\ifCLASSINFOpdf
\else
\fi
\begin{document}
%
\title{Fast Search on Binary Codes\\ by Weighted Hamming Distance}
%
%
%

\author{Zhenyu~Weng,~\IEEEmembership{Member,~IEEE,}
	Yuesheng~Zhu,~\IEEEmembership{Senior Member,~IEEE,}
	and~Ruixin~Liu
	\thanks{The authors are with Communication and Information Security Laboratory, Shenzhen Graduate School, Peking University, Shenzhen, China.
		E-mail: \{wzytumbler, zhuys, anne\_xin\}@pku.edu.cn. (Corresponding author: Yuesheng Zhu)}
}

%
%

\markboth{IEEE TRANSACTIONS ON Multimedia,~Vol.XX, No.XX, August~2021}%
{Shell \MakeLowercase{\textit{et al.}}: Bare Demo of IEEEtran.cls for IEEE Journals}
%



\maketitle

\begin{abstract}
Weighted Hamming distance, as a similarity measure between binary codes and binary queries, provides superior accuracy in search tasks than Hamming distance. However, how to efficiently and accurately find $K$ binary codes that have the smallest weighted Hamming distance to the query remains an open issue. In this paper, a fast search algorithm is proposed to perform the non-exhaustive search for $K$ nearest binary codes by weighted Hamming distance. By using binary codes as direct bucket indices in a hash table, the search algorithm generates a sequence to probe the buckets based on the independence characteristic of the weights for each bit. Furthermore, a fast search framework based on the proposed search algorithm is designed to solve the problem of long binary codes. Specifically, long binary codes are split into substrings and multiple hash tables are built on them. Then, the search algorithm probes the buckets to obtain candidates according to the generated substring indices, and a merging algorithm is proposed to find the nearest binary codes by merging the candidates. Theoretical analysis and experimental results demonstrate that the search algorithm improves the search accuracy compared to other non-exhaustive algorithms and provides orders-of-magnitude faster search than the linear scan baseline.
\end{abstract}

\begin{IEEEkeywords}
Binary codes, weighted Hamming distance, nearest neighbor search.
\end{IEEEkeywords}

%
\IEEEpeerreviewmaketitle

\section{Introduction}
%
%
%
%
\IEEEPARstart{W}{ith} the explosive growth of data in recent years, increasing demands for enhancing computation efficiency and reducing storage cost encourage research on binary codes. For example, binary image descriptors such as BRIEF~\cite{Calonder2012BRIEF}, ORB~\cite{6126544}, BRISK~\cite{6126542} and other binary image descriptors~\cite{6247715,8206261,7882718,8356135,7936534} are designed to represent image data, and then successfully used in various image applications, including image matching, 3D reconstruction and object recognition. Compared to the established descriptors such as
SIFT~\cite{lowe2004distinctive} or other descriptors~\cite{7887720} learned from convolutional networks, these lightweight binary descriptors are more compact to store and faster to compare. In addition to binary image descriptors, another notable application of binary codes is binary hashing~\cite{8477143,7990335,lin2020fast,lin2018supervised,weng2020concatenation,weng2019fast,DBLP:conf/aaai/LiuDML17,9126268,8656487,7377105,8247210,9046296,Lin2020,8543225,8705282,7915742,7870632}. By encoding high-dimensional data with binary codes and performing fast Hamming distance, hashing methods can implement the approximate nearest neighbor search efficiently with a small storage space. Moreover, the community has witnessed considerable promise of hashing in various search tasks such as clustering~\cite{gong2015web}, object detection~\cite{DBLP:conf/bmvc/KehlTNIL15}, image retrieval~\cite{DBLP:conf/aaai/LinJLSWW19,liu2018dense,7873258}, cross-modal retrieval~\cite{8423193} and person re-identification~\cite{7185403,7903578}.

\begin{figure}[t]
	\centering
	\includegraphics[width=0.45\textwidth]{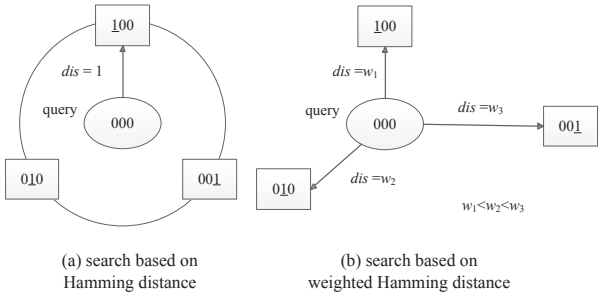}
	\caption{The comparison between Hamming distance and weighted Hamming distance.}
	\label{fig:label1}

\end{figure}

Despite the progress, there exists an ambiguity problem that different binary codes may share the same Hamming distance to the binary query point during Hamming distance comparison. To alleviate this problem, recent studies~\cite{Fan2013Learning,gordo2014asymmetric,zhang2013binary,Duan2015Weighted,8288679} propose to assign bitwise weights to each bit of binary codes, based on which the distance is measured by weighted Hamming distance. As shown in Fig.~\ref{fig:label1}, since there are different weights at each bit position between the query and the binary codes, the binary codes that share the same Hamming distance to the query yield different weighted Hamming distance to the query. By alleviating the ambiguity problem and broadening the comparison distance range, weighted Hamming distance can effectively improve the performance of binary codes while still enjoying small storage benefit. 

Although comparing binary codes by weighted Hamming distance can alleviate the ambiguity problem, the comparison is slower than that of Hamming distance. To accelerate the search process by weighted Hamming distance, some methods~\cite{gordo2014asymmetric,weng2016asymmetric} use lookup tables to compute the query-independent values in advance, so that the number of the computations in the search process can be reduced. However, it is still an exhaustive linear scan, which barricades their applications in the large-scale databases. To further accelerate the search process, some methods~\cite{Duan2015Weighted,norouzi2014fast} suggest that the search in weighted Hamming space can be accelerated by performing the non-exhaustive search in Hamming space at first and then re-ranking the candidates by weighted Hamming distance. However, it cannot provide the exact nearest neighbor search by weighted Hamming distance as the nearest neighbor may be discarded during the Hamming distance comparison, which results in the inferior accuracy. Therefore, how to efficiently and accurately find $K$ binary codes that have the smallest weighted Hamming distance to the query is still an open issue.

In this paper, we propose a fast search algorithm to perform the non-exhaustive search for $K$ nearest binary codes by weighted Hamming distance. By using a hash table populated with binary codes, the search algorithm generates a probing sequence of bucket indices based on the independence characteristic of the weights for each bit. It is faster than generating a probing sequence by directly sorting the bucket indices according to their weighted Hamming distance to the query. We prove that weighted Hamming distance between the query and the sequence of the generated bucket indices will increase monotonically. Furthermore, since using a hash table for long binary codes is not effective, we propose a search framework for long binary codes based on the proposed search algorithm. In this framework, long binary codes are split into substrings on which multiple hash tables are built, and the proposed search algorithm iteratively probes the table buckets until the query's nearest neighbors are found. In detail, by dividing the binary codes into disjoint substrings as the indices of the table buckets, the search algorithm generates a proper probing sequence for each substring. Then, the buckets are probed to obtain the candidates according to the probing sequences, and a merging algorithm is proposed to merge the candidates until $K$ nearest neighbors to the query are found. Also, we design a single multi-index hash table to reduce its practical storage cost. The results of extensive experiments demonstrate that the proposed search algorithm improves the search accuracy compared to other non-exhaustive search algorithms and is orders of magnitude faster than the linear scan baseline.

The rest of this paper is organized as follows: We briefly review related works in Section II. The search algorithm for compact codes by weighted Hamming distance and the search framework for long codes are presented in Section III and Section IV, respectively. The experimental results are presented in Section V. And finally, we conclude this paper in Section VI.

\section{Related Work}
\subsection{Weighted Hamming Distance}
In the large-scale approximate nearest neighbor search, encoding high-dimensional feature vectors with the compact binary codes offers two benefits: data compression and search efficiency. It is considered in~\cite{gordo2014asymmetric} that binarizing both query and database vectors is not strictly a requirement. It still enjoys the same two benefits but can provide superior accuracy by binarizing the database vectors and learning the weights for each bit of binary codes according to the original query information. The literatures~\cite{Duan2015Weighted,weng2016asymmetric,zhang2013binary,8288679,7516672} also improve the search accuracy by learning weights for the binary codes and replacing Hamming distance with weighted Hamming distance. They learn the weights from different respects, such as the types of the hash functions and the learning strategies. In addition, the weights for the binary image descriptors are learned in~\cite{Huang2017RWBD,Fan2013Learning} to improve their descriptive ability.

Although various methods about using weighted Hamming distance to improve search accuracy are developed, the methods about accelerating the search in weighted Hamming space are rare. The methods~\cite{Duan2015Weighted,norouzi2014fast} that accelerate the search by Hamming distance cannot be directly applied on weighted Hamming distance due to the essential difference between Hamming distance and weighted Hamming distance.

\subsection{Fast Search by Different Similarities}
Hamming distance is a commonly used similarity measure for comparing two binary codes, which can be computed efficiently by performing an XOR operation and then counting the number of ones in the result (computed by a single instruction, \_\_popcnt in C++). However, the search time of the linear scan on large scale datasets can still take several minutes~\cite{7780589}. To address this issue, MIH~\cite{norouzi2014fast,8463534} performs the non-exhaustive search by using a hash table that is populated with binary codes where each code is treated as an index in the hash table. To deal with long binary codes, MIH adopts multi-index hash tables that build multiple hash tables on disjoint binary code substrings and use each substring as an index of the corresponding hash table.

In addition to Hamming distance, there are other similarity measures that are used in different applications~\cite{matsui2015pqtable,8340865}. However, MIH cannot be directly used for other similarities to accelerate the search due to the essential differences between similarities. Therefore, to accelerate the search by the cosine similarity, AMIH~\cite{8340865} is developed to perform a non-exhaustive search on binary codes by using the connection between the Hamming distance
and the cosine similarity. Furthermore, AMIH also adopt multiple hash tables on binary substrings to deal with long binary codes. Product Quantization (PQ) codes~\cite{jegou2011product} are another kind of binary codes by using codebooks. PQ decomposes the high-dimensional vectors into the Cartesian product of sub-vectors and then quantizes these sub-vectors into binary codes by using codebooks. To accelerate the search for PQ, PQTable~\cite{matsui2015pqtable} develops a search method based on codebooks to perform a non-exhautive search. However, as described in~\cite{8340865}, how to perform the fast search for other measures of similarity such as weighted Hamming distance is still open.

In this paper, we propose a fast search algorithm to perform a non-exhaustive but exact $K$NN search on binary codes by weighted Hamming distance. A preliminary work was presented in the conference~\cite{AAAI2020query}. Apart from more detailed description, this paper differs from the conference version in the following aspects: 1) We provide a detailed theoretical analysis on the proposed search algorithm, including computational complexity. 2) We design a single multi-index hash table to replace multiple hash tables on substrings, which can effectively reduce the practical storage cost when being applied on the large-scale datasets. 3) We systematically describe the search framework to deal with long binary codes. For each module of the framework, detailed theoretical analysis and comparison with related works are given. 4) Much more experiments are conducted to demonstrate the effectiveness and efficiency of the search algorithm.

\section{Search by Weighted Hamming Distance}
\subsection{Definitions and Problem Statement}
Given the query $\mathbf{q}\in{\left \{ 0,1 \right \}}^{b}$ and the dataset $G=\left \{ \mathbf{g}_{1},\ldots\mathbf{g}_{N}\right \}$ where $\mathbf{g}_{i}\in{\left \{ 0,1 \right \}}^{b}$, Nearest Neighbor (NN) search (also the 1NN problem) aims to find the most closest item to \textbf{q} in $G$, which can be formulated as:
\begin{equation}
	NN(\mathbf{q})=\mathop{\arg\min}_{\mathbf{g}\in{G}}dis(\mathbf{q},\mathbf{g}),
\end{equation}
where $dis(\cdot,\cdot)$ is the distance between two items.

The $K$ Nearest Neighbor ($K$NN) search problem is the generalization of 1NN, aiming to find the $K$ closest items to the query. The focus of this paper is to efficiently solve the the $K$NN problem where $dis(\cdot,\cdot)$ is the weighted Hamming distance.

The weighted Hamming distance between the query $\mathbf{q}$ and the binary code $\bold{g}$ is defined as:
\begin{equation}
	d_{w}(\mathbf{q},\mathbf{g})=\sum_{i=1}^{b}w_{i}(q_{i}\oplus \mathrm{g}_{i}),
	\label{eqn2}
\end{equation}
where $\oplus$ is an XOR operation, $w_{i}:\left \{ 0,1 \right \}\rightarrow \mathbb{R}$ is a weight function for the $i^{th}$ bit, $q_{i}$ is the $i^{th}$ bit of $\mathbf{q}$, and $\mathrm{g}_{i}$ is the $i^{th}$ bit of $\mathbf{g}$.

\subsection{Algorithm Description}

Given the binary codes in $G$, we populate a hash table where each binary code is treated as a direct index of the table bucket. Then, we aim to find the $K$ closest binary codes that have the smallest weighted Hamming distance to the query $\mathbf{q}$.

Based on the independence characteristic of the weights for bits, we propose a fast search algorithm to generate a sequence of binary codes (bucket indices), where the weighted Hamming distance between the query and the sequence of binary codes will increase monotonically.

Since the query is fixed in the comparison between the query and each binary code, the weight values for the XOR result between the binary codes and the query can be pre-computed and stored. Hence, Eqn. (\ref{eqn2}) is rewritten as:
\begin{equation}
	d_{w}(\mathbf{g})=\sum_{i=1}^{b}\hat{w}_{i}(\mathrm{g}_{i}),
	\label{eqn3}
\end{equation}
where $\mathrm{g}_{i}$ is the $i^{th}$ bit of $\mathbf{g}$, $\hat{w}_{i}:\left \{ 0,1 \right \}\rightarrow \mathbb{R}$ is a function to store the pre-computed weight value for the $i^{th}$ bit and is defined as:
\begin{equation}
	\left \{
	\begin{array}{c}
		\hat{w}_{i}(0)={w}_{i}(0\oplus q_{i})
		\\
		\,\hat{w}_{i}(1)={w}_{i}(1\oplus q_{i}).
	\end{array}
	\right .
\end{equation}

The input values of the function $\hat{w}_{i}(\cdot )$ are either 0 or 1, which results in two possible values of $\hat{w}_{i}(\cdot)$. To construct an initial $b$-bit binary code $\mathbf{h}=\left [ h_1,\ldots, h_b \right ]$ that has the smallest weighted Hamming distance (smallest sum of weights) with the query, each bit $h_i$ of $\bold{h}$ is obtained as:
\begin{equation}
	h_i=\left\{
	\begin{array}{cc}
		0 & \hat{w}_i(0)\leq \hat{w}_i(1)
		\\
		1 & otherwise.
	\end{array}
	\right.
	\label{eqn5}
\end{equation}

When the $i^{th}$ bit of $\mathbf{h}$ changes ($i.e.$, from 0 to 1 or from 1 to 0), we use $\bar{h_i}$ to denote the changed $i^{th}$ bit. In this case, the weight for this bit will increase. The increased weight $\bigtriangleup \hat{w_i}$ of the $i^{th}$ bit is defined as:
\begin{equation}
	\bigtriangleup \hat{w_i}=\hat{w_i}(\bar{h_i})-\hat{w_i}(h_i).
	\label{eqn6}
\end{equation}

We first rank bits in an increasing order according to $\bigtriangleup \hat{w_i}$ in advance, and construct the initial smallest binary code $\mathbf{h}$. Then, we maintain a priority queue to generate a sequence of binary codes where the weighted Hamming distance between the binary codes and the query increases monotonically. The top of the priority queue is the binary code that has the smallest sum of weights in the priority queue. $\mathbf{h}$ is the first one that is pushed into the priority queue. When taking out the top binary code $\mathbf{\tilde{h}}$ from the priority queue and probing the corresponding hash bucket, two new binary codes are constructed from $\mathbf{\tilde{h}}$ by \textbf{two different operations} and pushed into the priority queue, respectively.

$\bold{Operation}$ $\bold{1}$ is to construct a binary code by changing the unchanged bit right next to the rightmost changed bit of $\bf{\tilde h}$ if the rightmost changed bit is not at the end of the current binary code. For example, given ${\bf{\tilde h}} = [h_1 \ldots \bar h_r \ldots h_b]$ where $\bar h_r$ is rightmost changed bit, the new binary code is constructed as ${\bf{\hat  h}} = [h_1 \ldots \bar h_r \bar h_{r+1} \ldots h_b]$.

$\bold{Operation}$ $\bold{2}$ is to construct a binary code by moving the rightmost changed bit of ${\bf{\tilde h}}$ to the next bit if the position of the rightmost changed bit is not at the end. For example, given ${\bf{\tilde h}} = [h_1 \ldots \bar h_r \ldots h_b]$ where $\bar h_r$ is rightmost changed bit, the new binary code is constructed as ${\bf{\dot  h}} = [h_1 \ldots h_r \bar h_{r+1} \ldots h_b]$.

\begin{figure}[t]
	\centering
	\includegraphics[width=0.5\textwidth]{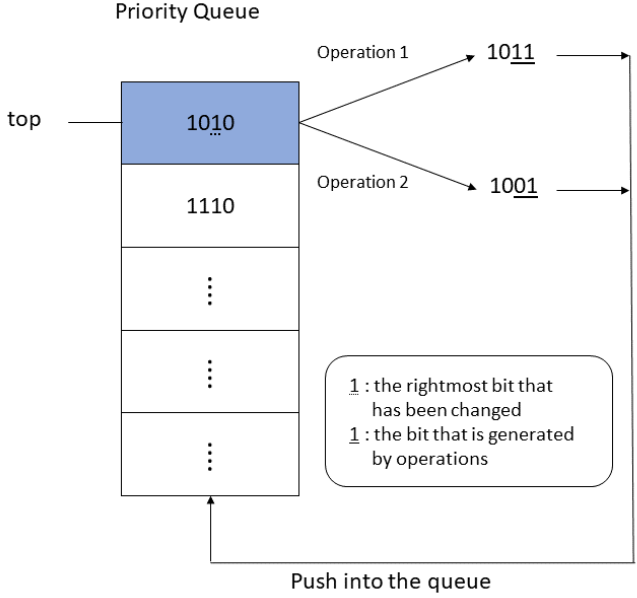}
	\caption{The procedure of generating binary codes by proposed operations. After popping up the top binary code 1010 as the bucket index to probe, two new codes 1011 and 1001 are generated by two operations, respectively.}
	\label{fig:label2}

\end{figure}

Hence, the fast search algorithm in weighted Hamming space is performed by continuously taking out the top binary code from the priority queue as the bucket index and generating new binary codes that are pushed into the priority queue. The search process is shown in Fig.~\ref{fig:label2}. After ranking the bits and generating the initial binary code according to the weight information and the query, two new binary codes are generated each time according to the above operations, resulting in a larger sum of weights than the current one. It should be noted that for the initial binary code $\bold{h}$ which has no changed bit, only $\bold{Operation}$ $\bold{1}$ is permitted, which means changing the first bit of the binary code.

\begin{algorithm}[t]
	\caption{The fast search algorithm by weighted Hamming distance}               
	\label{alg1}                      
	\begin{algorithmic}[1]
		\Require $\bold{q}$, $K$, $table$, $\bold{w}$              
		\Ensure $u$    \Comment{a set of ranked identifiers}   
		\State $u \leftarrow \emptyset$
		\State $pri\_que \leftarrow \emptyset$    \Comment{priority queue}
		\State $[pri\_que, order] \leftarrow$ Init($\bold{q}$, $\bold{w}$)
		\While {$|u| < K$}
		\State $code \leftarrow pri\_que.$top()
		\State $pri\_que.$pop()    \Comment{remove top item from queue}
		\State $pri\_que$.push(Operation1($code$, $order$))
		\State $pri\_que$.push(Operation2($code$, $order$))
		\State $ {\hat u}  \leftarrow table.$bucket($code$) \Comment{identifiers in the bucket}
		\State $u.$extend($ {\hat u} $)
		\EndWhile
	\end{algorithmic}
\end{algorithm}

The pseudocode for the proposed algorithm is shown in Alg. 1. Init() is a function that constructs the binary code $\bold{h}$ which has the smallest sum of weights according to the query $\bold{q}$ and the weights $\bold{w}$, and generates an order that denotes the positions of bits from smallest to largest according to Eqn. (\ref{eqn6}). Operation1() and Operation2() are two functions corresponding to above two operations to generate the new binary codes, respectively.

\subsection{Theoretical Analysis}

To prove that our algorithm can always find the binary code that has the smallest sum of weights among the un-probed binary codes, we begin with the following corollary.

\textbf{Corollary 1:} Every binary code can be generated by above two operations.
\begin{proof}
	It can be proved by mathematical induction.
	
	$\bf{Basis:}$ We have the binary code ${\bf{h}}_0$ = $\bf{h}$ which has no changed bits initially. Then by definition, ${\bf{\hat h}}_1$ is generated by changing the first bit of ${\bf{h}}_0$ using the first operation. It is easy to find that every binary code ${\bf{h}}_1$ which have 1 changed bit can be generated from ${\bf{\hat h}}_1$ according to the second operation.
	
	$\bf{Inductive}$ $\bold{step:}$ Assume every binary code ${\bf{h}}_z$, which has $z$ changed bits, can be generated. For every binary code ${\bf{h}}_{z+1} = [h_1 \ldots \bar h_i \ldots \bar h_j \ldots h_b]$ which has $z+1$ changed bits, where the $i^{th}$ bit and the $j^{th}$ bit are the $z^{th}$ and $(z+1)^{th}$ changed bits, respectively. It can be generated by the second operation from another binary code ${\bf{\hat h}}_{z+1} = [h_1 \ldots \bar h_i \bar h_{i+1} \ldots h_b]$, where the changed status of the $j^{th}$ bit is moved to the $(i+1)^{th}$ bit. Then ${\bf{\hat h}}_{z+1}$ can be generated by the first operation from the binary code ${\bf{\tilde h}}_z = [h_1 \ldots \bar h_i h_{i+1} \ldots h_j \ldots h_b]$, where the $(i+1)^{th}$ bit is changed back to the previous status and ${\bf{\tilde h}}_z$ has $z$ changed bits. Thereby, every binary code ${\bf{h}}_{z+1}$, which has $z+1$ changed bits, can be generated by two operations from $\bold{h}_z$.	
\end{proof}

Then, we prove the correctness of our algorithm by \textbf{Proposition 1}.

\textbf{Proposition 1:} The binary code that has the smallest sum of weights among the un-probed binary codes is always in the priority queue.
\begin{proof}
	It can be proved by mathematical induction.
	
	$\bf{Basis:}$ By definition, we have the binary code $\bf{h}$ which has no changed bits initially. It is the smallest among all the binary codes and is pushed into the priority queue.
	
	$\bf{Inductive}$ $\bold{step:}$ Assume the top item ${\bf{h}}_a$ of the priority queue is the smallest among the current un-probed binary codes. When it is taken out, two new binary codes are constructed and pushed into the priority queue. Assume there is another binary code ${\bf{h}}_b$ which is not in the priority queue and is smallest among the current un-probed binary codes after ${\bf{h}}_a$ is taken out. According to $\bold{Corollary}$ $\bold{1}$, ${\bf{h}}_b$ can be directly generated from ${\bf{h}}_c$ by the above two operations. ${\bf{h}}_c$ is smaller than ${\bf{h}}_b$. Since ${\bf{h}}_b$ is smallest among the current un-probed binary codes, ${\bf{h}}_c$ should have been probed. If ${\bf{h}}_c$ is probed and ${\bf{h}}_b$ is generated from ${\bf{h}}_c$, ${\bf{h}}_b$ should have been pushed into the priority queue. Obviously, the assumption is invalid. Therefore, the binary code which has smallest sum of weights among the un-probed binary codes is always in the priority queue.
	
\end{proof}

The sequencial procedure of generating the bucket indices can be regarded as the multiple sequences combination problem (one bit represents one sequence). An algorithm in~\cite{6915715,matsui2015pqtable} is used to solve the multiple sequences combination problem. However, this algorithm is not suitable in this situation. The algorithm in~\cite{6915715} can only traverse a few sequences ($e.g.$, 2 or 4) simultaneously to find the combination composing the bucket index that has the smallest weighted Hamming distance to the query. In this situation, there are $b$ sequences where $b$ is much larger than 4 such that the traversal space is very large.

\begin{figure*}[t]
	\centering
	\includegraphics[width=0.75\textwidth]{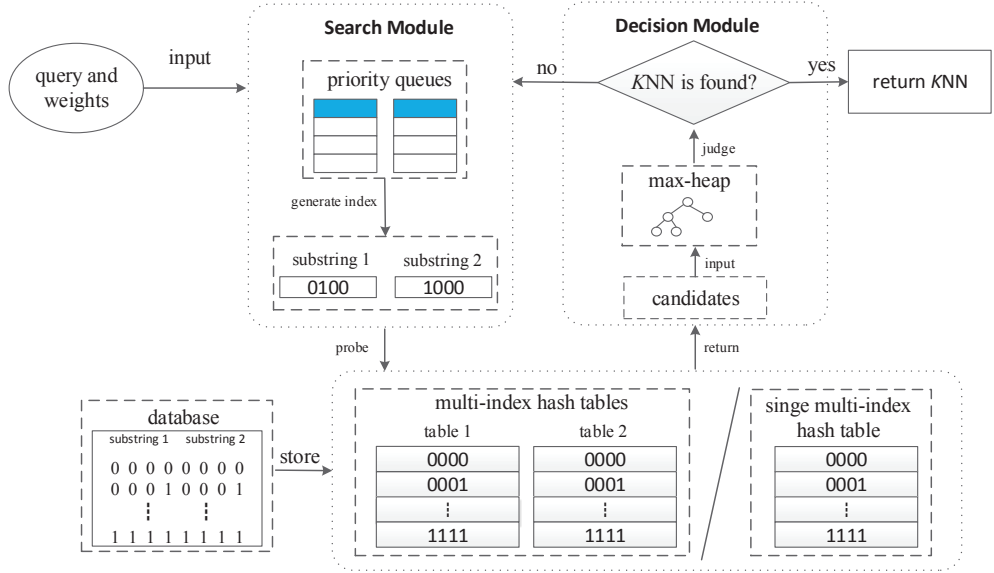}
	\caption{The search framework for long binary codes. The database binary codes are stored by applying multiple hash tables or a single multi-index hash table. Given the query and weights, the search module generates probing sequences for each substring, and the decision module merges the candidates from the probed buckets as well as determines whether the nearest neighbors are found.}
	\label{fig:label3}

\end{figure*}

\textbf{Complexity:} The complexity of the search algorithm to find nearest binary codes by weighted Hamming distance is $O((t+t^2)log(2t))$ where $t$ is the number of the probed buckets. 

\begin{proof}
	According to Alg. 1, at each step, when a new binary code is popping out from the priority queue to probe the bucket, at most two new binary codes are generated and put into the queue in $O(1)$ operations. Therefore, at the $t^{th}$ step of Alg. 1, when $t$ nearest binary codes have been output ($i.e.$, $t$ buckets are probed), the size of the priority queue is not larger than $2t$. As the operation of popping out an element in the priority queue is $O(zlog z)$ where $z$ is the number of elements in the queue, the complexity of probing $t$ buckets is
	\begin{equation}
		\begin{split}
			T&= O(2log2)+O(4log4)+\cdots+O((2t)log(2t)) \\
			&\le O(2log(2t))+\cdots+O(2tlog(2t)) \\
			&= O((t+t^2)log(2t)).  \\
		\end{split}
	\end{equation}

\end{proof}

\section{Search Framework for Long Codes}
In order to reach the desired search performance, hashing is usually constructed with a long code length (\emph{e.g.}, 64 bits). In this case, the computational cost is extremely heavy provided that a full hash table is adopted. As detailed in~\cite{norouzi2014fast}, the range of table index is exponentially expanded with the increase of code length. As a consequence, many table buckets become empty without any data item. Traversing these empty table buckets is redundant, and thus results in additional computational cost.

To mitigate this issue, MIH~\cite{norouzi2014fast} merges the buckets over different dimensions of the binary codes. Specifically, instead of creating one huge hash table, MIH separates the binary codes into $m$ smaller disjoint substrings, based on which a total of $m$ smaller hash tables are created to separately index their corresponding substrings. These hash tables are called multi-index hash tables. Then, MIH solves $m$ instances of the $K$NN search problem with Hamming distance to find the query's nearest neighbor in Hamming space, one per each hash table. More details can be found in~\cite{norouzi2014fast}. However, MIH is particularly designed for $K$NN search problem by Hamming distance, and thus fails to be directly adopted for weighted Hamming distance. This inspires us to further improve MIH so as to support the search by weighted Hamming distance.

\subsection{Search with Multi-Index Hash Tables}
In the proposed search framework, by separating the binary codes into $m$ smaller disjoint substrings, $m$ different hash tables are built. Each substring is then indexed within its corresponding hash table. The length of each substring is $\left \lceil b/m \right \rceil$ or $\left \lfloor b/m \right \rfloor$. For convenience, we assume that $m$ is a factor of $b$, and that the substrings comprise continuous bits.

Correspondly, we define $f:\left \{ 0,1 \right \}^{b/m}\rightarrow \mathbb{R}$ as a function to calculate the sum of weights of the substring $\mathbf{s}\in \left \{ 0,1 \right \}^{b/m}$, which is formulated as:

\begin{equation}
	f(\mathbf{s})=\sum_{j=1}^{b/m}\hat{w_j}(s_j),
\end{equation}
where $s_j$ is the $j^{th}$ bit of $\mathbf{s}$ and $\hat{w_j}:\left \{ 0,1 \right \}\rightarrow \mathbb{R}$ is the weight function for the $j^{th}$ bit.

As shown in Fig.~\ref{fig:label3}, the search framework is composed of two modules: the search module and the decision module. Given a query, it is similarly partitioned into $m$ substrings. Then, the search module provides a proper probing sequence according to each query substring information and the weight information. After probing some buckets and taking the candidates from the probed buckets, the decision module determines if the query's $K$ nearest neighbors are found or more buckets should be probed. Hence, the search framework iteratively probes the buckets until the $K$ nearest neighbors are found. In the following, we will describe these two modules in detail.

In the search module, each hash table maintains a priority queue according to the sum of weights of the corresponding substring. Each priority queue operates as in Alg. 1. Hence, a proper probing sequence is generated by taking out the top substring of every priority queue for each round. By treating the substring as the index of the hash table bucket in the corresponding table, the identifiers in the bucket are taken out as the candidates.

In the decision module, to compare the candidates from the bucket and determine if the query's nearest neighbors are found, a merging algorithm is proposed. In the algorithm, a $K$-size max-heap is built to filter the candidates. The root node of the max-heap has the largest sum of weights in the heap. Assume the node $\bold{r}\in \left \{ 0,1 \right \}^{b}$ in the max-heap is in the form of $\bold{r}=\left [ \bold{r}_1,\ldots,\bold{r}_m \right ]$ where $\bold{r}_i$ is the substring of $\bold{r}$ in the $i^{th}$ table. A function $g:\left \{ 0,1 \right \}^{b}\rightarrow \mathbb{R}$ to calculate the sum of weights of the node is defined as:
\begin{equation}
	{g(\bold{r})}=\sum_{i=1}^m f(\bold{r}_i).
	\label{eqn8}
\end{equation}

For each round, after the candidates are taken from each table, they are compared to the root node in the max-heap. If a candidate $\hat{\bold{r}}$ has a smaller sum of weights than the root node ($i.e.$, ${g(\hat{\bold{r}})<g(\bold{r})}$), the root node is thrown away and the candidate is inserted into the max-heap.

The criterion to decide if the query's nearest neighbors are found is that the root node of the max-heap is smaller or equal to a predefined threshold, which is:
\begin{equation}
	{g(\bold{r})}\leq S,
	\label{eqn9}
\end{equation}
where $S$ is the threshold.

In our implementation, we set $S$ as the sum of the weights of the current top substring of each priority queue. In detail, assume there are $m$ tables and a $b$-bit binary code $\bold{h}$ is partitioned into disjoint substrings $\bold{s}$. When the top substring $\bold{s}_i$ of the $i^{th}$ priority queue is taken out as the index of the bucket to be probed, the queue will have the new top substring $\hat{\bold{s}_{i}}$. The associated identifiers from the table bucket $\bold{s}_i$ of the $i^{th}$ table are taken out and compared with the root node of the max-heap. After probing the corresponding buckets in each table, the threshold $S$ is set as the the sum of weights from the current top substring of each priority queue, which is:
\begin{equation}
	S=\sum_{i=1}^m f(\hat{\bold{s}_{i}}).
	\label{eqn10}
\end{equation}

Further, to accelerate the search by meeting the threshold faster, we do not need to make a decision after probing the buckets in all hash tables for each round. Instead, if the decision criterion is met after probing the first $j$ tables' buckets, the $K$ nearest neighbors are found and the search process can terminate. The threshold for probing the first $j$ tables is calculated as:
\begin{equation}
	\bar{S}= \sum_{i=1}^j f(\hat{\mathbf{s_i}})+\sum_{i=j+1}^m f(\mathbf{s_i}),
	\label{eqn11}
\end{equation}
where $\hat{\mathbf{s_i}}$ is the top substring of the priority queue in the current round and $\mathbf{s_i}$ is the top substring of the priority queue in the last round.

Obviously, $S \leq \bar{S}$. Therefore, we have
\begin{equation}
	{g(\bold{r})}\leq S \leq \bar{S}.
\end{equation}

Hence, using Eqn. (\ref{eqn11}) as the threshold can make the search procedure terminate earlier (meet the threshold earlier) than using Eqn. (\ref{eqn10}) as the threshold.

In what follows, we will prove the correctness of the decision criterion.

\textbf{Proposition 2:} The binary codes that are found and stored in the max-heap have the smallest sum of weights among all binary codes.

\begin{proof}
	This can be proved by contradiction. Assume there exists an identifier that is not found yet and its corresponding binary code is $\bold{\tilde r}$. Its sum of weights $g(\bold{\tilde r})$ (Eqn. (\ref{eqn8})) is smaller than that of the root node $\bold{r}$ in the max-heap ($i.e.$ $g(\bold{\tilde r}) < g(\bold{r})$). Since $g({\bf{r}}) \le \bar{S}$ (Eqn. (\ref{eqn11})) and $g({\bf{\tilde r}}) < g({\bf{r}}),$ $g({\bf{\tilde r}}) < \bar{S}$. Then, at least one of the disjoint substrings of $\bold{\tilde r}$ is smaller than the top substring of the corresponding priority queue. This substring should have been taken out from the queue according to \textbf{Proposition 1}. Since the substring is taken out as a bucket to be probed, $\bold{\tilde r}$ should have been inserted into the max-heap. Obviously the assumption is invalid. Therefore, the binary codes that are found and stored in the max-heap have the smallest sum of weights among all binary codes.
\end{proof}

\begin{algorithm}[t]
	\caption{The fast search framework for long codes}               
	\label{alg2}                      
	\begin{algorithmic}[1]
		\Require $\bold{q}, table[], K, m$, $\bold{w}$              
		\Ensure $max\_heap$     
		\State $max\_heap \leftarrow \emptyset$
		\State $[pri\_que[], order[]] \leftarrow$ Split(Init($\bold{q}$,$\bold{w}$),$m$)
		\While {$!max\_heap.$satisfied()}
		\For {$i \leftarrow$ $1$ to $m$}
		\State $code[i] \leftarrow pri\_que[i].$top()
		\State $pri\_que[i].$pop()
		\State $pri\_que[i].$push(Operation1($code[i], order[i]$))
		\State $pri\_que[i].$push(Operation2($code[i], order[i]$))
		\State $max\_heap.$insert($table[i].$hash($code[i]$))
		\If {$max\_heap.$satisfied()}
		\State    break
		\EndIf
		\EndFor
		\EndWhile
	\end{algorithmic}
\end{algorithm}

\begin{table*}[htbp]
	\centering
	\caption{The percentage ($\%$) of the usage of the bucket indices on GIST1M and SIFT1B.}
	\begin{tabular}{|c|c|c|c|c|}  \hline
		\toprule	
		bit	& \multicolumn{2}{c|}{GIST1M}     & \multicolumn{2}{c|}{SIFT1B} \\   \cmidrule{2-5}
		&  $|T_{one}|/|T_{all}|$ &  $|T_{all}-T_{one}|/|T_{all}|$ &  $|T_{one}|/|T_{all}|$ & $|T_{all}-T_{one}|/|T_{all}|$  	\\ \midrule
		32  & 6.0  &  94.0 & 100.0 & 0.0 \\ \midrule
		64  & 0.0  &  100.0 & 96.0 & 4.0 \\ \midrule
		128  & 0.0  &  100.0 & 91.0 & 9.0 \\ \midrule

	\end{tabular}%
	\label{tab:addlabel0}%
\end{table*}%

The pseudocode for the search framework is shown in Alg. 2. Init(), Operation1() and Operation2() are the same functions as in the Alg. 1. $m$ denotes the number of substrings for the binary code. $table[], pri\_que[]$, and $order[]$ denote a set of tables, a set of priority queues, and a set of bit rankings for each substring, respectively. Split() is a function that splits the binary code and the weights into $m$ parts. $max\_heap$.satisfied() denotes whether the root node of the max-heap satisfies the stopping criterion according to Eqn. (\ref{eqn9}) and Eqn. (\ref{eqn11}).

In our preliminary work~\cite{AAAI2020query}, we want to find an order of probing hash tables so that the root node of the max-heap be smaller or equal to $S$ faster. To this end, for each round, before probing the buckets, we obtain an order by sorting the tables according to the difference $\Delta {f_i}$ from smallest to largest. The difference is calculated as:
\begin{equation}
	\Delta {f_i} = f(\hat{\mathbf{s}_i}) - f(\mathbf{s}_i),
\end{equation}
where $\hat{\mathbf{s}_i}$ is the top substring of the priority queue in the current round and $\mathbf{s}_i$ is the top substring of the priority queue in the last round.

Hence, the decision criterion in~\cite{AAAI2020query} is:
\begin{equation}
	\hat{S} = \sum_{i=1}^j f(\hat{\mathbf{s}}_{order[i]})+\sum_{i=j+1}^m f(\mathbf{s}_{order[i]}),
	\label{eqn15}
\end{equation}
where the order is obtained by sorting the tables according to the difference $\Delta {f_i}$.

In Sec. V, we will show that although using the threshold of Eqn. (\ref{eqn15}) can reduce the number of probed buckets, it adds the time cost of sorting the hash tables and makes the search slower than using the threshold of Eqn. (\ref{eqn11}).

PQTable~\cite{matsui2015pqtable} considers another criterion to determine whether the query's nearest neighbors are found, which can be presented in the form of Eqn. (\ref{eqn9}). In PQTable, the search process terminates when finding a binary code that has appeared $m$ times and there are $K-1$ binary codes that yield smaller (or equal) weighted Hamming distance to the query than that binary code. Obviously, when a binary code appears $m$ times, it means that its corresponding substrings are all probed. Hence, the threshold is the sum of the weights of the probed binary substring in each table. In detail, if the top substrings from the first $j$ priority queues are taken out, the threshold is formulated as:
\begin{equation}
	\tilde{S} = \sum_{i=1}^j f(\mathbf{s}_i) + \sum_{i=j+1}^m f(\tilde{\mathbf{s}_i}),
\end{equation}
where $\mathbf{s}_i$ is the taken substring in the current round and $\tilde{\mathbf{s}_i}$ is the taken substring in the last round.

Obviously, $\tilde{S}<\bar{S}$ for the same round since $\tilde{\mathbf{s}_i}<\mathbf{s}_i<\hat{\mathbf{s}_i}$. Therefore, the search procedure by PQTable will ends later then the search according to Eqn. (\ref{eqn11}), resulting in more candidates to be compared and larger search time cost.

\subsection{Search with a Single Multi-Index Hash Table}
\label{singleMulti-indexHT}
As analyzed in~\cite{norouzi2014fast,matsui2015pqtable}, the practical storage cost of multi-index hash tables is larger than the theoretical storage cost, due to the null pointers in the empty buckets.

To reduce the empty buckets and the practical storage cost, we show that multi-index hash tables can be merged into one hash table over the bucket indices. In other words, the binary codes are still separated into $m$ disjoint substring and each substring is the bucket index within the same hash table rather than multiple hash tables.

For multi-index hash tables, let $T_i$ denote the set of the non-empty bucket indices in the $i^{th}$ hash table, $T_{one}$ denote the set of the non-empty bucket indices that exist in only one of $m$ hash tables, and $T_{all}=T_1\cup T_2\cup \ldots \cup T_m$ denote the set of the non-empty bucket indices in $m$ hash tables. It is easy to know that the percentage of the non-empty bucket indices that exist in only one of $m$ hash tables over the non-empty bucket indices in $m$ hash tables is represented as $|T_{one}|/|T_{all}|$, and the percentage of the non-empty bucket indices that exist in more than one hash tables over the non-empty bucket indices in $m$ hash tables is represented as $|T_{all}-T_{one}|/|T_{all}|$. Here, $|\cdot|$ is the cardinality of the set. Table~\ref{tab:addlabel0} shows the statistics about $|T_{one}|/|T_{all}|$ and $|T_{all}-T_{one}|/|T_{all}|$ on the GIST1M dataset and the SIFT1B dataset~\cite{jegou2011product}. The data in the dataset is hashed to binary codes by Locality-Sensitive Hashing (LSH)~\cite{andoni2006near}. Following ~\cite{norouzi2014fast,8340865,matsui2015pqtable}, the value of $m$ ($i.e.$, the number of the substrings) is set to $b/log_2 n$, where $n$ is the data size and $b$ is the length of the binary code. Hence, on GIST1M, the number of the substrings is 2 for 32 bits, 4 for 64 bits and 8 for 128 bits ($i.e.$, the length of the substring is 16 bits). On SIFT1B, the number of the substrings is 1 for 32 bits, 2 for 64 bits and 4 for 128 bits ($i.e.$, the length of the substring is 32 bits). According to the results in Table~\ref{tab:addlabel0}, most bucket indices are shared by more than one table on GIST1M while few buckets indices are shared by more than one table on SIFT1B. Although merging the tables may make the buckets include more candidates that are not the query's nearest neighbors, resulting in the additional computational cost, this effect rarely happens on SIFT1B as only few buckets indices are shared by more than one table.

Hence, instead of creating $m$ hash tables where there are many empty buckets, we desgin a single multi-index hash table where the binary codes are still separated into $m$ disjoint substring and each substring is indexed within the same hash table. The single multi-index hash table can be directly used by the proposed search framework on the binary substrings, as shown in Fig.~\ref{fig:label3}.

\section{Experiments}
\subsection{Datasets and Settings}
In the experiments, we will measure the efficiency and effectiveness of the proposed search algorithm on the binary codes in the weighted Hamming space. For convenience, the binary codes in the weighted Hamming space are also called the weighted binary codes. Various kinds of weighted hashing methods are used to generate different weighted binary codes for two datasets: GIST1M~\cite{jegou2011product} and SIFT1B~\cite{jegou2011product}.

The GIST1M dataset contains 1 million 960-dimensional GIST descriptors~\cite{oliva2001modeling} which are global descriptors, and extracted from Tiny image set~\cite{torralba200880}. The dataset contains 1,000 queries.

The SIFT1B dataset contains 1 billion 128-dimensional SIFT descriptors~\cite{lowe2004distinctive}, which are local descriptors. The dataset contains 10,000 queries. GIST and SIFT are often used in computer vision areas, such as feature matching and image retrieval.

Two kinds of hashing methods are used to encode the feature vectors with the binary codes. Locality-Sensitive Hashing (LSH)~\cite{andoni2006near} is used to generate data-independent hash functions, which map the feature vectors to the binary codes, and ~\cite{gordo2014asymmetric} is used to learn the weights for the binary codes. ITerative Quantization (ITQ)~\cite{gong2013iterative} is used to generate data-dependent hash functions, and ~\cite{gordo2014asymmetric} is used to generate the weights.

\subsection{Comparison with Different Bits}
We compare the proposed search algorithm, Fast Search on Weighted Binary Codes (FSWBC), with Linear Scan, MIH\_binary, and MIH\_weight. Linear Scan denotes the exhaustive search for the weighted binary codes and is accelerated by adopting the lookup tables~\cite{gordo2014asymmetric}. MIH\_binary denotes the multi-index hashing algorithm~\cite{norouzi2014fast} that performs the non-exhaustive search for the binary codes according to Hamming distance. MIH\_weight denotes the multi-index hashing~\cite{norouzi2014fast} that performs the non-exhaustive search for the binary codes according to Hamming distance and then selects the neighbors among the candidates according to weighted Hamming distance. These algorithms are all implemented in C++. Following ~\cite{norouzi2014fast}, the value of $m$ (the number of the substrings) for the search algorithms is set to $b/log_2 n$, except for linear scan. All the experiments are run on a single core Intel Core-i5-6500 CPU with 64GB of memory.\footnote{All source codes are available on \url{https://github.com/zyweng2pku/FSWBC}}

To measure the search accuracy for the $K$NN search, pre@$K$ for Weighted Binary Codes (WBC) is used to measure the search accuracy for WBC. It is defined as the fraction of the true retrieved weighted binary codes to the retrieved weighted binary codes and formulated as follows
\begin{equation}
	{\rm pre@{\emph K}   \,\, for\,\, WBC=\frac{the\ true\ retrieved\ WBC}{\emph K}}.
\end{equation}

In addition to the search accuracy, we measure the search efficiency by the search time cost and the speed-up factor compared to the linear scan for the weighted binary codes. The speed-up factor is defined as dividing the run-time cost of the linear scan by the run-time cost of the test algorithm, which is formulated as follows
\begin{equation}
	\rm{speed-up\ factor}=\frac{time\ of\ linear\ scan}{time\ of\ test\ method}.
\end{equation}

\begin{table}[htbp]
	\centering
	\caption{The search accuracy comparison of pre@1(\%) on GIST1M with LSH.}
	
	\begin{tabular}{|p{6em}|c|c|c|c|c|}
		\toprule
		\multicolumn{1}{|c|}{} & 8 bits & 16 bits & 32 bits & 64 bits & 128 bits \\
		\midrule
		Linear Scan & 100   & 100   & 100   & 100   & 100 \\
		\midrule
		MIH\_binary & 100   & 99.6  & 35.6  & 30.3  & 30.5 \\
		\midrule
		MIH\_weight & 100   & 99.6  & 38.7  & 37.8  & 44.7 \\
		\midrule
		\textbf{FSWBC} & 100   & 100   & 100   & 100   & 100 \\
		\bottomrule
	\end{tabular}%
	\label{tab:addlabel1}%
\end{table}%

\begin{table}[htbp]
	\centering
	\caption{The search accuracy comparison of pre@1(\%) on GIST1M with ITQ.}
	\begin{tabular}{|p{6em}|c|c|c|c|c|}
		\toprule
		\multicolumn{1}{|c|}{} & 8 bits & 16 bits & 32 bits & 64 bits & 128 bits \\
		\midrule
		Linear Scan & 100   & 100   & 100   & 100   & 100 \\
		\midrule
		MIH\_binary & 100   & 99    & 45    & 32.2  & 36 \\
		\midrule
		MIH\_weight & 100   & 99    & 47.2  & 38.1  & 49 \\
		\midrule
		\textbf{FSWBC} & 100   & 100   & 100   & 100   & 100 \\
		\bottomrule
	\end{tabular}%
	\label{tab:addlabel2}%
\end{table}%

\begin{table}[htbp]
	\centering
	\caption{The search accuracy comparison of pre@1(\%) on SIFT1B with LSH.}
	\begin{tabular}{|p{6em}|c|c|c|c|c|}
		\toprule
		\multicolumn{1}{|c|}{} & 8 bits & 16 bits & 32 bits & 64 bits & 128 bits \\
		\midrule
		Linear Scan & 100   & 100   & 100   & 100   & 100 \\
		\midrule
		MIH\_binary & 100   & 100   & 88.6  & 35.8  & 34.4 \\
		\midrule
		MIH\_weight & 100   & 100   & 88.6  & 39.6  & 44.5 \\
		\midrule
		\textbf{FSWBC} & 100   & 100   & 100   & 100   & 100 \\
		\bottomrule
	\end{tabular}%
	\label{tab:addlabel3}%
\end{table}%

\begin{table*}[htbp]
	\centering
	\caption{The search efficiency comparison on GIST1M with LSH for different bits.}
	\begin{tabular}{|p{5.2em}|c|c|c|c|c|c|c|c|c|c|}
		\toprule
		\multicolumn{1}{|c|}{\multirow{3}[4]{*}{}} & \multicolumn{2}{c|}{8 bits} & \multicolumn{2}{c|}{16 bits} & \multicolumn{2}{c|}{32 bits} & \multicolumn{2}{c|}{64 bits} & \multicolumn{2}{c|}{128 bits} \\
		\cmidrule{2-11}    \multicolumn{1}{|c|}{} & \multicolumn{1}{p{4.215em}|}{Time} & \multicolumn{1}{p{4.215em}|}{Speed } & \multicolumn{1}{p{4.215em}|}{Time} & \multicolumn{1}{p{4.215em}|}{Speed } & \multicolumn{1}{p{4.215em}|}{Time} & \multicolumn{1}{p{4.215em}|}{Speed } & \multicolumn{1}{p{4.215em}|}{Time} & \multicolumn{1}{p{4.215em}|}{Speed } & \multicolumn{1}{p{4.215em}|}{Time} & \multicolumn{1}{p{4.215em}|}{Speed } \\
		\multicolumn{1}{|c|}{} & \multicolumn{1}{p{4.215em}|}{(ms)} & \multicolumn{1}{p{4.215em}|}{up} & \multicolumn{1}{p{4.215em}|}{(ms)} & \multicolumn{1}{p{4.215em}|}{up} & \multicolumn{1}{p{4.215em}|}{(ms)} & \multicolumn{1}{p{4.215em}|}{up} & \multicolumn{1}{p{4.215em}|}{(ms)} & \multicolumn{1}{p{4.215em}|}{up} & \multicolumn{1}{p{4.215em}|}{(ms)} & \multicolumn{1}{p{4.215em}|}{up} \\
		\midrule
		Linear Scan & 10.01 & 1     & 15.33 & 1     & 25.4  & 1     & 44.52 & 1     & 89.04 & 1 \\
		\midrule
		MIH\_binary & 0.02  & 500.5 & 0.03  & 511   & 0.08  & 317.5 & 0.84  & 53    & 7.94  & 11.2 \\
		\midrule
		MIH\_weight & 0.02  & 500.5 & 0.03  & 511   & 0.08  & 317.5 & 0.84  & 53    & 7.94  & 11.2 \\
		\midrule
		\textbf{FSWBC} & 0.02  & 500.5 & 0.03  & 511   & 0.11  & 230.9 & 2.11  & 21.1  & 18.37 & 4.8 \\
		\bottomrule
	\end{tabular}%

	\label{tab:addlabel4}%
\end{table*}%

\begin{table*}[htbp]
	\centering
	\caption{The search efficiency comparison on GIST1M with ITQ for different bits.}
	\begin{tabular}{|p{5.2em}|c|c|c|c|c|c|c|c|c|c|}
		\toprule
		\multicolumn{1}{|c|}{\multirow{3}[4]{*}{}} & \multicolumn{2}{c|}{8 bits} & \multicolumn{2}{c|}{16 bits} & \multicolumn{2}{c|}{32 bits} & \multicolumn{2}{c|}{64 bits} & \multicolumn{2}{c|}{128 bits} \\
		\cmidrule{2-11}    \multicolumn{1}{|c|}{} & \multicolumn{1}{p{4.215em}|}{Time} & \multicolumn{1}{p{4.215em}|}{Speed } & \multicolumn{1}{p{4.215em}|}{Time} & \multicolumn{1}{p{4.215em}|}{Speed } & \multicolumn{1}{p{4.215em}|}{Time} & \multicolumn{1}{p{4.215em}|}{Speed } & \multicolumn{1}{p{4.215em}|}{Time} & \multicolumn{1}{p{4.215em}|}{Speed } & \multicolumn{1}{p{4.215em}|}{Time} & \multicolumn{1}{p{4.215em}|}{Speed } \\
		\multicolumn{1}{|c|}{} & \multicolumn{1}{p{4.215em}|}{(ms)} & \multicolumn{1}{p{4.215em}|}{up} & \multicolumn{1}{p{4.215em}|}{(ms)} & \multicolumn{1}{p{4.215em}|}{up} & \multicolumn{1}{p{4.215em}|}{(ms)} & \multicolumn{1}{p{4.215em}|}{up} & \multicolumn{1}{p{4.215em}|}{(ms)} & \multicolumn{1}{p{4.215em}|}{up} & \multicolumn{1}{p{4.215em}|}{(ms)} & \multicolumn{1}{p{4.215em}|}{up} \\
		\midrule
		Linear Scan & 9.56  & 1     & 15.21 & 1     & 25.35 & 1     & 44.47 & 1     & 89.04 & 1 \\
		\midrule
		MIH\_binary & 0.02  & 478   & 0.03  & 507   & 0.13  & 195   & 0.69  & 64.4  & 6.47  & 13.8 \\
		\midrule
		MIH\_weight & 0.02  & 478   & 0.03  & 507   & 0.13  & 195   & 0.69  & 64.4  & 6.47  & 13.8 \\
		\midrule
		\textbf{FSWBC} & 0.02  & 478   & 0.03  & 507   & 0.08  & 316.9 & 1.33  & 33.4  & 13.78 & 6.5 \\
		\bottomrule
	\end{tabular}%
	\label{tab:addlabel5}%
\end{table*}%

\begin{table*}[htbp]
	\centering
	\caption{The search efficiency comparison on SIFT1B with LSH for different bits.}
	\begin{tabular}{|p{5.2em}|c|c|c|c|c|c|c|c|c|c|}
		\toprule
		\multicolumn{1}{|c|}{\multirow{3}[4]{*}{}} & \multicolumn{2}{c|}{8 bits} & \multicolumn{2}{c|}{16 bits} & \multicolumn{2}{c|}{32 bits} & \multicolumn{2}{c|}{64 bits} & \multicolumn{2}{c|}{128 bits} \\
		\cmidrule{2-11}    \multicolumn{1}{|c|}{} & \multicolumn{1}{p{4.215em}|}{Time} & \multicolumn{1}{p{4.215em}|}{Speed } & \multicolumn{1}{p{4.215em}|}{Time} & \multicolumn{1}{p{4.215em}|}{Speed } & \multicolumn{1}{p{4.215em}|}{Time} & \multicolumn{1}{p{4.215em}|}{Speed } & \multicolumn{1}{p{4.215em}|}{Time} & \multicolumn{1}{p{4.215em}|}{Speed } & \multicolumn{1}{p{4.215em}|}{Time} & \multicolumn{1}{p{4.215em}|}{Speed } \\
		\multicolumn{1}{|c|}{} & \multicolumn{1}{p{4.215em}|}{(ms)} & \multicolumn{1}{p{4.215em}|}{up} & \multicolumn{1}{p{4.215em}|}{(ms)} & \multicolumn{1}{p{4.215em}|}{up} & \multicolumn{1}{p{4.215em}|}{(ms)} & \multicolumn{1}{p{4.215em}|}{up} & \multicolumn{1}{p{4.215em}|}{(ms)} & \multicolumn{1}{p{4.215em}|}{up} & \multicolumn{1}{p{4.215em}|}{(ms)} & \multicolumn{1}{p{4.215em}|}{up} \\
		\midrule
		Linear Scan & 10154.23 & 1     & 15684.52 & 1     & 25701.79 & 1     & 45117.23 & 1     & 90930.07 & 1 \\
		\midrule
		MIH\_binary & 8.69  & 1168.5 & 8.81  & 1780.3 & 8.65  & 2971.3 & 10.05 & 4489.3 & 46.75 & 1945.0 \\
		\midrule
		MIH\_weight & 8.69  & 1168.5 & 8.81  & 1780.3 & 8.65  & 2971.3 & 10.14 & 4449.4 & 54.84 & 1658.1 \\
		\midrule
		\textbf{FSWBC} & 8.8   & 1153.9 & 8.82  & 1778.3 & 8.34  & 3081.7 & 10.2  & 4423.3 & 67.8  & 1341.2 \\
		\bottomrule
	\end{tabular}%
	\label{tab:addlabe6}%
\end{table*}%

Table~\ref{tab:addlabel1} and Table~\ref{tab:addlabel2} show the search accuracy comparison of finding the nearest neighbor (1-NN) on the weighted binary codes generated by LSH and ITQ on GIST1M, respectively. Table~\ref{tab:addlabel3} shows the search accuracy comparison on SIFT1B. The length of binary codes range from 8 bits to 128 bits. MIH\_binary and MIH\_weight can achieve 100\% accuracy for 8 bits, while their accuracy decreases as the number of bits increases. When the number of bits is small, there are only a few buckets in the table so that the nearest neighbor can be easily found in the corresponding bucket indexed by the query. As the number of bits increases, there are more buckets in the table and the bucket indexed by the query may be empty. Hence, MIH\_binary and MIH\_weight need to probe the other buckets that are not the nearest to the query by weighted Hamming distance, resulting in the decreased accuracy. Instead, FSWBC and Linear Scan can achieve 100\% accuracy for different bits. It demonstrates that FSWBC can perform the exact NN search on weighted binary codes while MIH\_binary and MIH\_weight cannot.

\begin{figure}[t]
	\centering
	\includegraphics[width=0.4\textwidth]{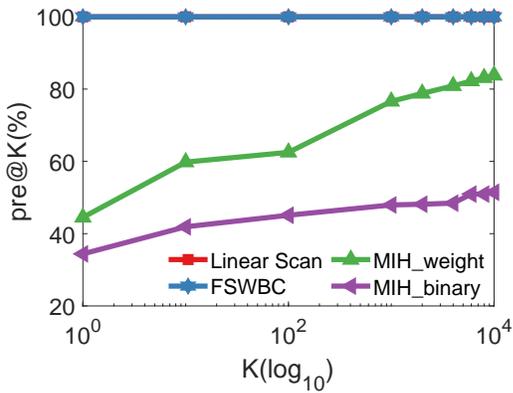}
	\caption{The search accuracy of different $K$ on SIFT1B.}
	\label{fig:label4}

\end{figure}

Fig.~\ref{fig:label4} shows the search accuracy comparison of $K$-NN search on SIFT1B for 128 bits. According to the results, with the number of returned neighbors increasing, MIH\_weight is obviously better than MIH\_binary as MIH\_weight performs the additional operation of selecting the neighbors among the candidates according to weighted Hamming distance compared to MIH\_binary. FSWBC can perform the exact $K$-NN search on weighted binary codes.

Table~\ref{tab:addlabel4} and Table~\ref{tab:addlabel5} show the search efficiency comparison of finding the nearest neighbor on the weighted binary codes generated by LSH and ITQ on GIST1M, respectively. Table~\ref{tab:addlabe6} shows the search efficiency comparison on SIFT1B. The length of binary codes range from 8 bits to 128 bits. The results in the tables show that MIH\_binary, MIH\_weight, and FSWBC can achieve a large speed-up compared to the linear scan baseline, especially on SIFT1B that includes one billion binary codes. Compared to MIH\_binary, MIH\_weight performs the additional operation of selecting the neighbors by weighted Hamming distance and thus achieves a better search accuracy for the weighted binary codes as shown in Table~\ref{tab:addlabel3}, but needs a larger time cost as shown in Table~\ref{tab:addlabe6}. In some cases, such as 32 bits on SIFT1B, FSWBC is faster than MIH\_binary. It is because FSWBC probes less buckets and less candidates than MIH\_binary, which will be analyzed later. According to the results, FSWBC can effectively accelerate the search on the weighted binary codes, especially on SIFT1B where it can achieve more than one thousand times speed-up compared to the linear scan baseline.

\subsection{Search Performance Analysis}

\begin{table*}[htbp]
	\centering
	\caption{The search performance analysis on GIST1M with ITQ.}
	\begin{tabular}{|c|c|c|c|c|c|c|c|c|c|c|c|}
		\toprule
		bit   & method & \multicolumn{3}{c|}{$K=$1} & \multicolumn{3}{c|}{$K=$10} & \multicolumn{3}{c|}{$K=$100} & Storage \\
		\cmidrule{3-11}          &       & Time  & \multicolumn{1}{c|}{buckets} & \multicolumn{1}{c|}{candidates} & Time  & \multicolumn{1}{c|}{buckets} & \multicolumn{1}{c|}{candidates} & Time  & \multicolumn{1}{c|}{buckets} & \multicolumn{1}{c|}{candidates} & (MB) \\
		&       & (ms)  &       &       & (ms)  &       &       & (ms)  &       &       &  \\
		\midrule
		32    & MIH\_binary & 0.13  & 20    & 4469  & 0.19  & 68    & 6327  & 0.38  & 217   & 11637 & 15.9 \\
		\cmidrule{2-2}          & PQTable & 1.66  & 26    & 5045  & 2.08  & 60    & 6900  & 3.16  & 161   & 11366 & 15.9 \\
		\cmidrule{2-2}          & \textbf{FSWBC\_single} & 0.09  & 12    & 4480  & 0.22  & 41    & 6458  & 0.63  & 138   & 12280 & 13.7 \\
		\cmidrule{2-2}          & \textbf{FSWBC} & 0.08  & 12    & 4380  & 0.19  & 41    & 5935  & 0.55  & 138   & 10478 & 15.9 \\
		\midrule
		64    & MIH\_binary & 0.69  & 288   & 14165 & 1.28  & 690   & 26549 & 2.44  & 1523  & 48289 & 27.6 \\
		\cmidrule{2-2}          & PQTable & 13.44 & 956   & 39404 & 18.15 & 1415  & 52291 & 26.90 & 2297  & 73728 & 27.6 \\
		\cmidrule{2-2}          & \textbf{FSWBC\_single} & 2.11  & 250   & 25360 & 4.84  & 607   & 52365 & 10.42 & 1378  & 102915 & 25.6 \\
		\cmidrule{2-2}          & \textbf{FSWBC} & 1.33  & 250   & 14727 & 3.02  & 607   & 27037 & 6.39  & 1378  & 48869 & 27.6 \\
		\midrule
		128   & MIH\_binary & 6.47  & 2230  & 67395 & 10.56 & 4163  & 106668 & 16.25 & 7302  & 159670 & 54.1 \\
		\cmidrule{2-2}          & PQTable & 130.20 & 12623 & 226645 & 154.70 & 15585 & 263330 & 187.49 & 19783 & 305136 & 54.1 \\
		\cmidrule{2-2}          & \textbf{FSWBC\_single} & 35.91 & 2172  & 231884 & 59.70 & 4140  & 361761 & 89.09 & 7405  & 498446 & 49.5 \\
		\cmidrule{2-2}          & \textbf{FSWBC} & 13.78 & 2172  & 70597 & 23.37 & 4140  & 112344 & 38.91 & 7405  & 166520 & 54.1 \\
		\bottomrule
	\end{tabular}%
	\label{tab:addlabel6}%
\end{table*}%

\begin{table*}[htbp]
	\centering
	\caption{The search performance analysis on SIFT1B with LSH.}
	\begin{tabular}{|c|c|c|c|c|c|c|c|c|c|c|c|}
		\toprule
		bit   & method & \multicolumn{3}{c|}{$K=$1} & \multicolumn{3}{c|}{$K=$10} & \multicolumn{3}{c|}{$K=$100} & Storage \\
		\cmidrule{3-11}          &       & Time  & buckets & candidates & Time  & buckets & candidates & Time  & buckets & candidates & (MB) \\
		&       & (ms)  &       &       & (ms)  &       &       & (ms)  &       &       &  \\
		\midrule
		32    & MIH\_binary & 8.65  & 8     & 2648  & 8.65  & 51    & 2697  & 8.78  & 317   & 3122  & 13,365.3 \\
		\cmidrule{2-2}          & PQTable & 8.48  & 2     & 1     & 8.48  & 6     & 10    & 8.62  & 38    & 100   & 13,365.3 \\
		\cmidrule{2-2}          & \textbf{FSWBC\_single} & 8.34  & 2     & 1     & 8.35  & 6     & 10    & 8.45  & 38    & 100   & 13,365.3 \\
		\cmidrule{2-2}          & \textbf{FSWBC} & 8.34  & 2     & 1     & 8.35  & 6     & 10    & 8.45  & 38    & 100   & 13,365.3 \\
		\midrule
		64    & MIH\_binary & 10.05 & 1725  & 9548  & 13.62 & 7125  & 26799 & 25.86 & 26671 & 80146 & 27,296.3 \\
		\cmidrule{2-2}          & PQTable & 20.34 & 1047  & 11982 & 33.74 & 2538  & 23744 & 75.56 & 7625  & 57155 & 27,296.3 \\
		\cmidrule{2-2}          & \textbf{FSWBC\_single} & 10.81 & 434   & 6219  & 16.3  & 1670  & 16862 & 38.73 & 6500  & 51340 & 24,319.0 \\
		\cmidrule{2-2}          & \textbf{FSWBC} & 10.2  & 434   & 6074  & 15.63 & 1670  & 14681 & 37.64 & 6500  & 47383 & 27,296.3 \\
		\midrule
		128   & MIH\_binary & 46.75 & 37475 & 149762 & 105.54 & 117339 & 367169 & 238.42 & 341004 & 845881 & 53,711.1 \\
		\cmidrule{2-2}          & PQTable & 617.31 & 62290 & 428009 & 955.91 & 99390 & 623251 & 1803.9 & 185956 & 1007101 & 53,711.1 \\
		\cmidrule{2-2}          & \textbf{FSWBC\_single} & 70.85 & 10136 & 114475 & 195.77 & 30762 & 278245 & 578.49 & 91726 & 658012 & 42,190.0 \\
		\cmidrule{2-2}          & \textbf{FSWBC} & 67.8  & 10136 & 106672 & 187.89 & 30762 & 255151 & 555.18 & 91726 & 589718 & 53,711.1 \\
		\bottomrule
	\end{tabular}%
	\label{tab:addlabel7}%
\end{table*}%

\begin{table*}[htbp]
	\centering
	\caption{The decision criterion comparison on SIFT1B.}
	\begin{tabular}{|c|c|c|c|c|c|c|c|c|c|c|}
		\toprule
		bit   & method & \multicolumn{3}{c|}{$K=$1} & \multicolumn{3}{c|}{$K=$10} & \multicolumn{3}{c|}{$K=$100} \\
		\cmidrule{3-11}          &       & Time  & buckets & candidates & Time  & buckets & candidates & Time  & buckets & candidates \\
		&       & (ms)  &       &       & (ms)  &       &       & (ms)  &       &  \\
		\midrule
		32    & \textbf{FSWBC} & 8.34  & 2     & 1     & 8.35  & 6     & 10    & 8.45  & 38    & 100 \\
		\cmidrule{2-2}          & \textbf{FSWBC\_sort} & 8.34  & 2     & 1     & 8.35  & 6     & 10    & 8.45  & 38    & 100 \\
		\cmidrule{2-2}          & \textbf{FSWBC\_PQ} & 8.34  & 2     & 1     & 8.35  & 6     & 10    & 8.45  & 38    & 100 \\
		\midrule
		64    & \textbf{FSWBC} & 10.2  & 434   & 6074  & 15.63 & 1670  & 14681 & 37.64 & 6500  & 47383 \\
		\cmidrule{2-2}          & \textbf{FSWBC\_sort} & 10.49 & 434   & 5715  & 15.95 & 1670  & 16089 & 38.05 & 6500  & 48988 \\
		\cmidrule{2-2}          & \textbf{FSWBC\_PQ} & 19.67 & 1047  & 11982 & 32.42 & 2538  & 23744 & 71.12 & 7625  & 57155 \\
		\midrule
		128   & \textbf{FSWBC} & 67.8  & 10136 & 106672 & 187.89 & 30762 & 255151 & 555.18 & 91726 & 589718 \\
		\cmidrule{2-2}          & \textbf{FSWBC\_sort} & 68.46 & 10137.24 & 106099.4 & 189.17 & 30763 & 254849 & 558.07 & 91726 & 589635 \\
		\cmidrule{2-2}          & \textbf{FSWBC\_PQ} & 551.29 & 62290.81 & 428009 & 676.01 & 99390 & 623251 & 1218.58 & 185957 & 1007101 \\
		\bottomrule
	\end{tabular}%
	\label{tab:addlabel8}%
\end{table*}%

Table~\ref{tab:addlabel6} and Table~\ref{tab:addlabel7} show the performance analysis of different search algorithms for the weighted binary codes learned by ITQ on GIST1M and learned by LSH on SIFT1B, respectively. The number of the probed buckets and the number of the compared candidates are two key factors that influence the search efficiency~\cite{norouzi2014fast}. FSWBC\_single denotes FSWBC with the proposed single multi-index hash table. In addition to MIH\_binary, we compare FSWBC with PQTable~\cite{matsui2015pqtable} which can perform the non-exhaustive search on weighted binary codes by translating the weighted binary codes in the form of PQ codes in advance. Specifically, the binary codes are split into disjoint parts each of which consists of continuous 8-bit binary codes. Then, each part can be regarded as a codebook, and PQTable is applied. As the performance of MIH\_binary is almost same as that of FSWBC for 8 bits and 16 bits, the performance analysis is focused on 32 bits, 64 bits, and 128 bits.

On GIST1M, FSWBC\_single is slower than FSWBC since FSWBC\_single probes the same number of the buckets as FSWBC does but compares more candidates. Although the Hamming distance computation is faster than the weighted Hamming distance computation, FSWBC is competitive to MIH\_binary for 32 bits since FSWBC probes less buckets and compares less candidates than MIH\_binary. Especially for returning 1NN on 32-bit binary codes, as FSWBC and FSWBC\_single probe 40\% less buckets than MIH\_binary, they are faster than MIH\_binary. PQTable is slower than FSWBC and FSWBC\_single from 32 bits to 128 bits as PQTable probes more buckets and candidates than FSWBC and FSWBC\_single. Compared to other algorithms, FSWBC\_single can reduce the practical storage cost as it adopts the single multi-index hash table.

On SIFT1B, FSWBC\_single is as fast as FSWBC for 32 bits since the length of the substring is 32 bits and they both use one hash table. Since FSWBC use one hash table for 32 bits, the number of candidates found in the buckets is the same as the number of required nearest neighbors. Also, we can see that with the number of buckets increasing, the time cost of FSWBC increases smaller, which confirms with the complexity of the proposed search algorithm in Sec. III.C. FSWBC\_single is a little slower than FSWBC for 64 bits and 128 bits as FSWBC\_single compares a little more candidates than FSWBC. FSWBC and FSWBC\_single are both competitive to MIH\_binary for 32 bits and 64 bits due to the less probed buckets and the less compared candidates than MIH\_binary. PQTable is slower than FSWBC and FSWBC\_single from 32 bits to 128 bits as PQTable probes more buckets and candidates than FSWBC and FSWBC\_single. In terms of the practical storage cost for the hash tables to store binary codes, using the single multi-index hash table can effectively reduce the storage cost than using the multi-index hash tables. According to the results in Table~\ref{tab:addlabel7}, by replacing the multi-index hash tables with the single multi-index hash table, almost 10\% and 20\% storage cost can be reduced for 64 bits and 128 bits, respectively. Hence, by using the single multi-index hash table, FSWBC\_single can effectively reduce the storage cost with little additional search cost compared to FSWBC on SIFT1B.

Table~\ref{tab:addlabel8} shows the comparison of different decision criterion in the decision module on SIFT1B. In the table, FSWBC\_sort denotes the decision criterion in~\cite{AAAI2020query} and FSWBC\_PQ denotes the decision criterion in PQTable ~\cite{matsui2015pqtable}. According to the results, FSWBC, FSWBC\_sort, and FSWBC\_PQ have the same time cost for 32 bits since the length of the substring is equal to the length of the binary code and thus they use one hash table. FSWBC\_sort and FSWBC are both faster than FSWBC\_PQ for 64 bits and 128 bits as FSWBC\_PQ probes more buckets and compares more candidates. In general, FSWBC\_sort probes less buckets and compareds less candidates than FSWBC. However, the reduction is not much and FSWBC\_sort needs addition computational cost for generating the probing sequence of the substrings. Therefore, FSWBC\_sort is slower than FSWBC. By comparing FSWBC\_PQ in Table~\ref{tab:addlabel8} with PQTable in Table~\ref{tab:addlabel7}, PQTable is slower than FSWBC with the same decision criterion, which demonstrates that FSWBC is more efficient in the bucket search procedure than PQTable as FSWBC is designed based on the characteristics of weighted binary codes.

\section{Conclusion}
In this paper, we focus on the $K$ nearest neighbor search problem on binary codes by weighted Hamming distance. By using a hash table populated with binary codes, a search algorithm is proposed to generate a probing sequence of binary codes where weighted Hamming distance between the query and the elements of the sequence will increase monotonically. Hence, the table buckets are probed one by one according to the probing sequence until the nearest binary codes to the query by weighted Hamming distance are found. Further, a search framework based on the search algorithm is proposed to deal with long binary codes. In this framework, multiple hash tables are built on binary code substrings and the search algorithm generates a probing sequence for each substring. The buckets in these tables are probed according to the sequences and the nearest neighbors are found by merging the candidates from the probed buckets according to the proposed merging algorithm. In addition, we design a single multi-index hash table, which can replace the multiple hash tables to reduce the practical storage cost with little extra time cost in the search process. The experiments show that the proposed search algorithm can improve the search accuracy compared to other search algorithms and achieve orders of magnitude speed up than the linear search baseline, especially when the size of the dataset is large. 

\bibliographystyle{IEEEtran}
\bibliography{egbib}

\vfill
%








\end{document}